\documentclass[journal]{IEEEtran} %
\IEEEoverridecommandlockouts
\usepackage{cite}
\usepackage{amsmath,amssymb,amsfonts}
\usepackage{algorithmic}
\usepackage{graphicx}
\usepackage{textcomp}
\usepackage{xcolor}
\usepackage{svg}
\usepackage{float}
\usepackage{subcaption} 
\usepackage{array} 
\usepackage{booktabs} 
\usepackage{multirow} 
\usepackage{colortbl}

\def\BibTeX{{\rm B\kern-.05em{\sc i\kern-.025em b}\kern-.08em
    T\kern-.1667em\lower.7ex\hbox{E}\kern-.125emX}}
\begin{document}

\title{LRSAA: Large-scale Remote Sensing Image Target Recognition and Automatic Annotation\\
% delete or comment-out the following line before submission
%{\footnotesize \textsuperscript{*}Note: Sub-titles are not captured in Xplore and should not be used}
%\thanks{Identify applicable funding agency here. If none, delete this.}
}

\author{%%%% author names
    \IEEEauthorblockN{1\textsuperscript{st} Wuzheng Dong }% first author
    \IEEEauthorblockN{2\textsuperscript{nd} Yujuan Zhu }% delete this line if not needed
     \IEEEauthorblockN{3\textsuperscript{rd} Sheng Zhang}% delete this line if not needed
    % duplicate the line above as many times as needed to list all authors
    \\%%%% author affiliations
    \IEEEauthorblockA{\textit{School of Mathematical Sciences, Nankai University, Tianjin, China}}\\% first affiliation
    %\IEEEauthorblockA{\textit{School of Economics, Nankai University, Tianjin, China}}\\% delete this line if not needed
    % duplicate the line above as many times as needed to list all affiliations
    %%%% corresponding author contact details
    \IEEEauthorblockA{{aerovane, 2210455}@mail.nankai.edu.cn}
}

\maketitle

\begin{abstract}
    This paper introduces a novel method for object recognition and automatic labeling in large-area remote sensing images, called LRSAA. The proposed method integrates the YOLOv11 and MobileNetV3-SSD object detection algorithms through ensemble learning to enhance overall model performance. Additionally, it utilizes Poisson disk sampling segmentation techniques along with the EIOU metric to optimize both the training and inference processes of segmented images, culminating in an integrated results framework. This approach not only minimizes computational resource requirements but also achieves an effective balance between accuracy and processing speed. The source code for this project is publicly accessible at https://github.com/anaerovane/LRSAA.
\end{abstract}


\section{Introduction}
Remote sensing target recognition encompasses the automatic detection and classification of terrestrial targets utilizing satellite or aerial imagery.  This technology plays a pivotal role in various domains, including environmental protection, resource management, disaster monitoring, and national defense.  Recent advancements in deep learning have significantly enhanced both the accuracy and efficiency of recognition processes, facilitating the detection of smaller targets and improving performance under complex conditions.

Nevertheless, several challenges persist within this technological landscape.  These include limited adoption of advanced models such as MobileNetV3-SSD and YOLOv11, reliance on single-model strategies that may compromise robustness, and inefficiencies associated with processing large-scale images.  To address these issues, we propose an innovative framework designed to enhance object recognition performance and automate labeling in extensive remote sensing images.

Our proposed method incorporates several pivotal technical innovations:

\begin{itemize}

\item \textbf{Ensemble Learning}: Utilizing MobileNetV3-SSD and YOLOv11, with the capability to be extended to multiple models.

\item \textbf{Enhanced Non-Maximum Suppression (NMS)}: Employing the EIOU metric to achieve improved suppression performance.

\item  \textbf{Poisson Disk Sampling}: For efficient dataset partitioning and accurate object recognition.

\end{itemize}

We conducted training and evaluation of our solution using the XView dataset as well as urban remote sensing images. The results indicate that our approach achieves substantial improvements in both accuracy and speed when compared to existing solutions. We assert that the LRSAA model has the potential to advance methodologies in remote sensing image analysis, thereby facilitating more effective and precise applications of geographic information systems.

\begin{figure}
    \centering \includegraphics[width=0.8\linewidth]{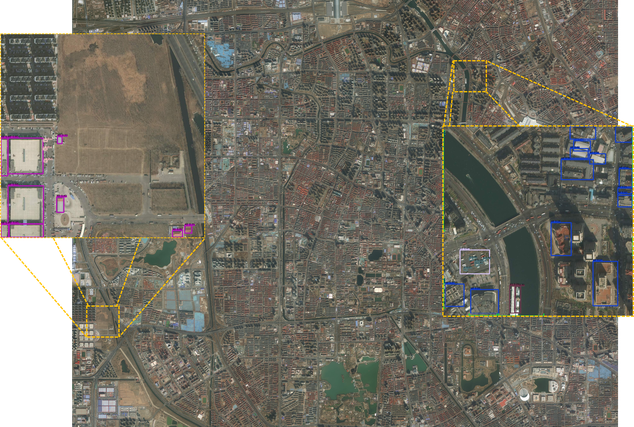}
    \caption{Practical application effects of the LRSAA model: a demonstration of detailed annotation results on Tianjin urban remote sensing images with 0.6m precision.} 
    \label{sample-figure}
\end{figure}

Through these technological advancements, we have trained and evaluated the Large-Scale Remote Sensing Automatic Annotation (LRSAA) model using the XView dataset. Subsequently, the model was applied to remote sensing images of Tianjin for automatic annotation. To further validate our approach, we incorporated a certain proportion of synthetic data, generated from automatically annotated images, into real remote sensing images. The results confirmed that the inclusion of synthetically annotated data can significantly enhance the recognition capabilities of the automatic annotation model. We anticipate that this work will contribute to the advancement of remote sensing image analysis, fostering more efficient and accurate acquisition and utilization of geographic information.

\section{Related Works}

\subsection{Object detection}

Object detection, a fundamental domain within computer vision, has experienced substantial advancements attributed to deep learning techniques and the availability of large annotated datasets. Initially dependent on hand-crafted features and traditional algorithms such as Support Vector Machines (SVMs)\cite{svm} and Histogram of Oriented Gradients (HOG)\cite{hog}, the field progressed significantly with the introduction of deep Convolutional Neural Networks (CNNs). Notable models including R-CNN, Fast R-CNN, and Faster R-CNN\cite{rcnn3} implemented a two-step approach for region proposal and classification, achieving high accuracy\cite{citing1}; however, this came at a considerable computational expense.

To enhance efficiency, single-shot detectors like Single Shot MultiBox Detector (SSD)\cite{ssd} were developed. These models utilize a VGG backbone to directly predict object classes and bounding boxes from feature maps. RetinaNet tackled class imbalance issues by employing a focal loss function, which improved detection performance for small and densely packed objects through Feature Pyramid Networks (FPN) that facilitate multi-scale feature extraction.

Recent developments in this area include YOLOv11 and MobileNetV3-SSD\cite{mbnet1,mbnet2}. MobileNet is recognized for its lightweight architecture that employs Depthwise Separable Convolutions to minimize computational load while reducing model parameters. Introduced in 2019, MobileNetV3 further optimizes these aspects using Neural Architecture Search (NAS) and NetAdapt methodologies to enhance both speed and accuracy specifically tailored for mobile devices\cite{mbnet3}. The combination of MobileNetV3 with SSD results in an efficient object detection framework characterized by reduced computational demands and storage requirements.\cite{mbnet4,mbnet5}

YOLO has gained acclaim for its rapid processing capabilities alongside impressive accuracy metrics\cite{yolo1,yolo2}; it conceptualizes object detection as a singular regression problem by predicting bounding boxes along with class probabilities directly from images. Released in September 2024, YOLOv11 demonstrates state-of-the-art performance across various tasks concerning accuracy, speed, and overall efficiency\cite{yolo7}.

In conclusion, the evolution of object detection has transitioned from conventional methods towards advanced deep learning frameworks; nevertheless, cutting-edge technologies continue to be underutilized within the realm of remote sensing\cite{yolo5,yolo6}.

\subsection{Poisson Disk Sampling}

Poisson Disk Sampling is a sophisticated sampling technique that generates sample points uniformly distributed in space, while ensuring that the minimum distance between any two points is at least a specified radius \( r \)\cite{pds1}. This method holds significant importance in the field of computer graphics, particularly in applications such as texture generation, object distribution, resampling, and rendering. In 2007, Bridson introduced an efficient algorithm\cite{pds2}, which has served as the foundation for numerous subsequent implementations and enhancements of Poisson disk sampling. The Poisson disk sampling algorithm continues to be optimized and its application scope expanded\cite{pds3,pds4}.

\section{Methods}

% \begin{figure*}[htbp]
%     \centering 
%     {\scriptsize
%     \includesvg[width=0.8\textwidth]{figure/fig8.svg}}
%     \caption{Methods:
% Our methodology comprises four stages: Step A, B, C, and D. In Step A, we utilize Poisson disk sampling and segmentation techniques to partition the large-scale image into smaller segments. Step B involves training YOLOv11 and MobileNetV3-SSD models on these smaller images to detect objects. For Step C, the detection results from the smaller images are mapped back onto the original large-scale image to maintain spatial consistency. Finally, in Step D, we augment the original dataset with synthetic data generated through this process, allowing for re-training the models with an enriched dataset that includes proportional synthetic samples. This approach ensures continuous improvement and adaptation of the models to diverse scenarios.} 
%     \label{sample-figure}
% \end{figure*}

\begin{figure*}
    \centering \includegraphics[width=0.8\linewidth]{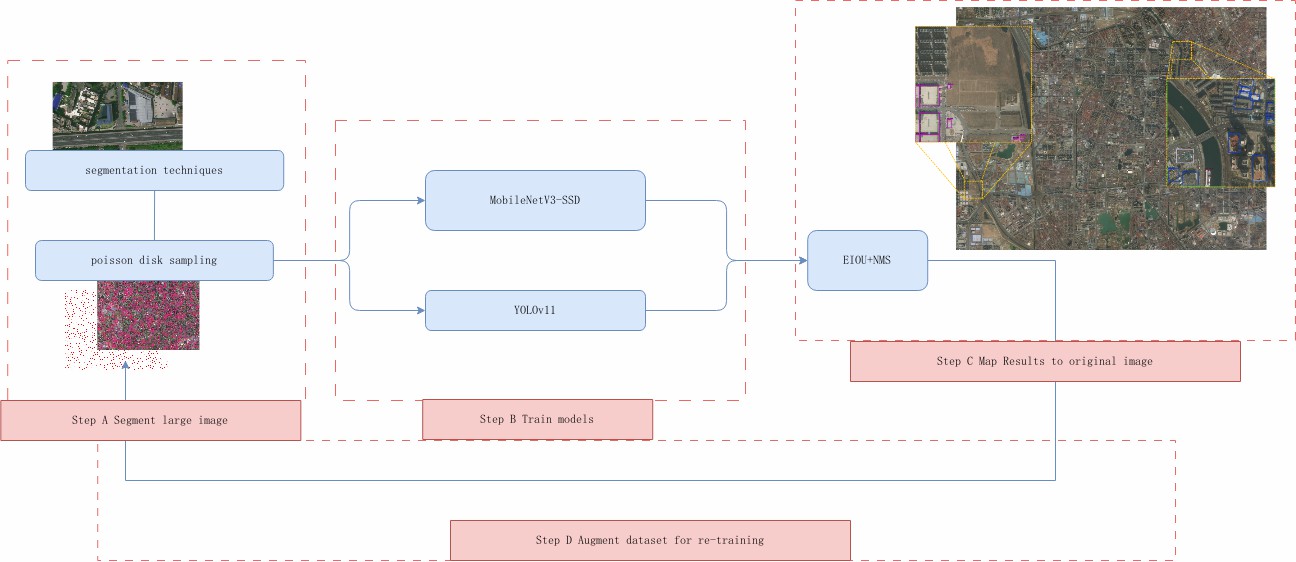}
    \caption{Methods:
Our methodology comprises four stages: Step A, B, C, and D. In Step A, we utilize Poisson disk sampling and segmentation techniques to partition the large-scale image into smaller segments. Step B involves training YOLOv11 and MobileNetV3-SSD models on these smaller images to detect objects. For Step C, the detection results from the smaller images are mapped back onto the original large-scale image to maintain spatial consistency. Finally, in Step D, we augment the original dataset with synthetic data generated through this process, allowing for re-training the models with an enriched dataset that includes proportional synthetic samples. This approach ensures continuous improvement and adaptation of the models to diverse scenarios.} 
    \label{sample-figure}
\end{figure*}

The methodological framework of this study can be summarized into four core steps:

\begin{itemize}

\item \textbf{1.} YOLOv11 and MobileNetV3-SSD trainging

\item \textbf{2.} Poisson disk sampling and segmentation

\item \textbf{3.} Map the results detected on the small segmented images back to the original large-scale image

\item \textbf{4.} Re-train with synthetic data added proportionally.

\end{itemize}

Below is a detailed description of the steps:

Initially, we compiled a comprehensive dataset comprising numerous labeled images for object detection and employed YOLOv11 and MobileNetV3-SSD models for preliminary model training. These models were selected due to their favorable balance of speed, accuracy, and versatility.

To enhance model generalization and robustness, we incorporated an image segmentation technique based on Poisson disk sampling, which uniformly distributes points across an image. This method is founded on the principles of the Poisson process and can be mathematically expressed as follows:

\[ \forall x_i, x_j \in S, \|x_i - x_j\| \geq r, i \neq j \]

We utilized Bridson's algorithm to accelerate the sampling process by defining both the sampling domain \(D\) and minimum distance \(r\), while employing a grid structure to improve efficiency. The procedure involves selecting points and verifying distances to ensure uniform distribution.

Subsequently, the segmented sub-images underwent object detection using the trained models. This required transforming coordinates to map detected bounding boxes back to their corresponding positions in the original image. This transformation is accomplished through the application of matrix operations:

\[ \begin{bmatrix} x_{min} \\ y_{min} \\ x_{max} \\ y_{max} \end{bmatrix} = \begin{bmatrix} 1 & 0 & 0 & 0 \\ 0 & 1 & 0 & 0 \\ 1 & 0 & 0 & 0 \\ 0 & 1 & 0 & 0 \end{bmatrix} \begin{bmatrix} x_{min}' \\ y_{min}' \\ x_{max}' \\ y_{max}' \end{bmatrix} + \begin{bmatrix} w \cdot n_x \\ h \cdot n_y \\ w \cdot n_x \\ h \cdot n_y \end{bmatrix} \]

\begin{figure}[H]
    \centering
    \begin{subfigure}{0.235\textwidth}
        \includegraphics[width=\linewidth]{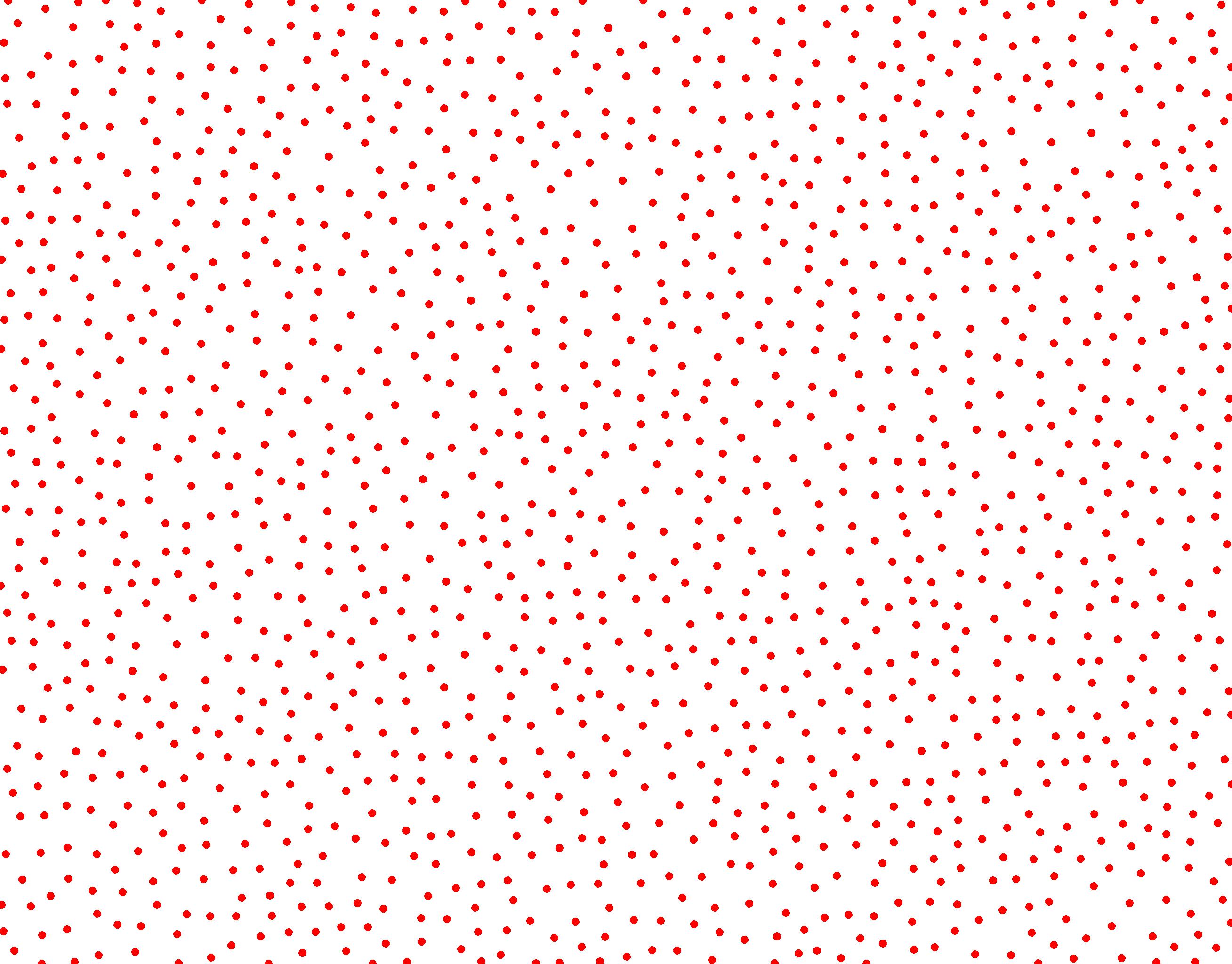}
    \end{subfigure}\hfill
    \begin{subfigure}{0.235\textwidth}
        \includegraphics[width=\linewidth]{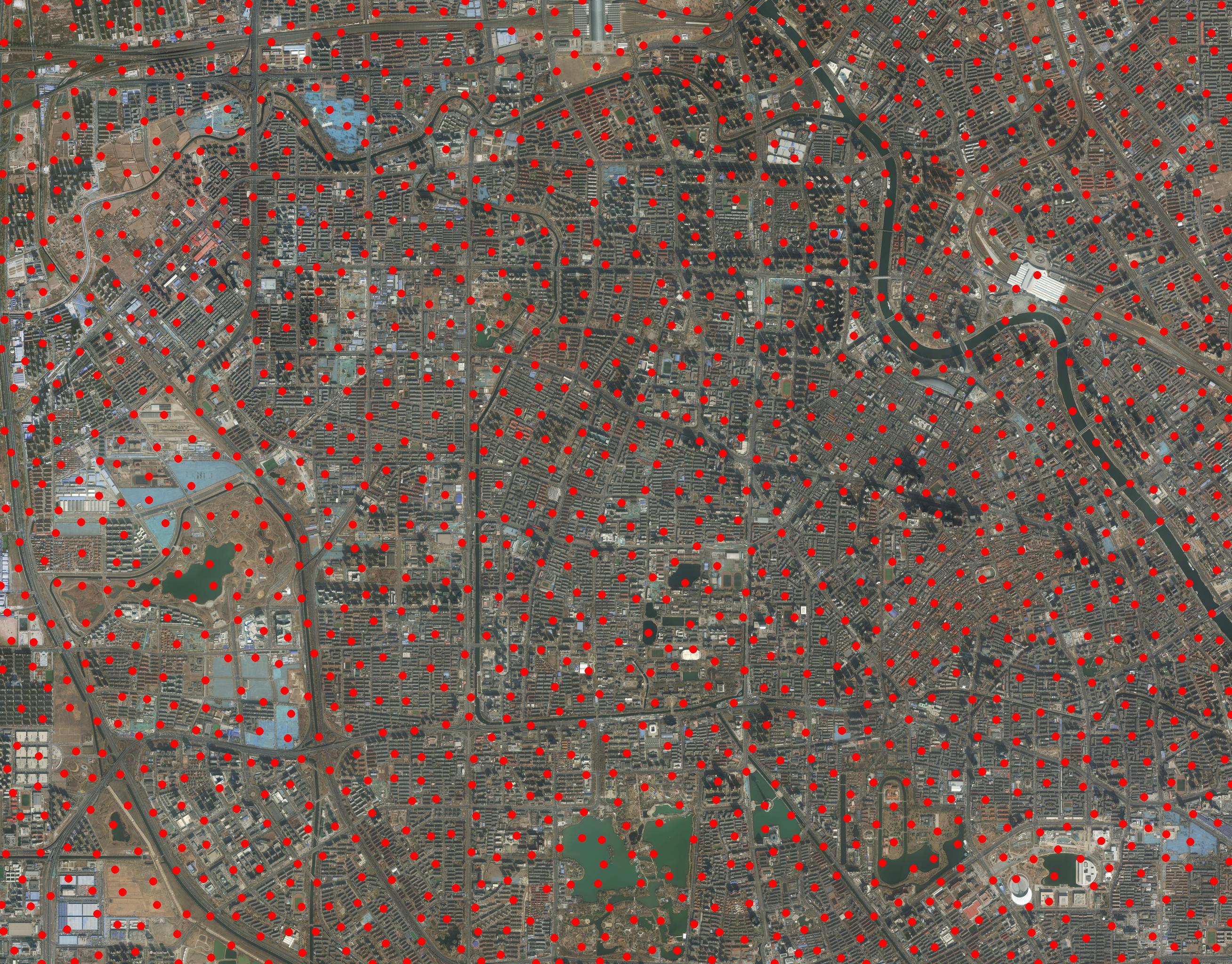}
    \end{subfigure}

    \vspace{1em} 

    \begin{subfigure}{0.235\textwidth}
        \includegraphics[width=\linewidth]{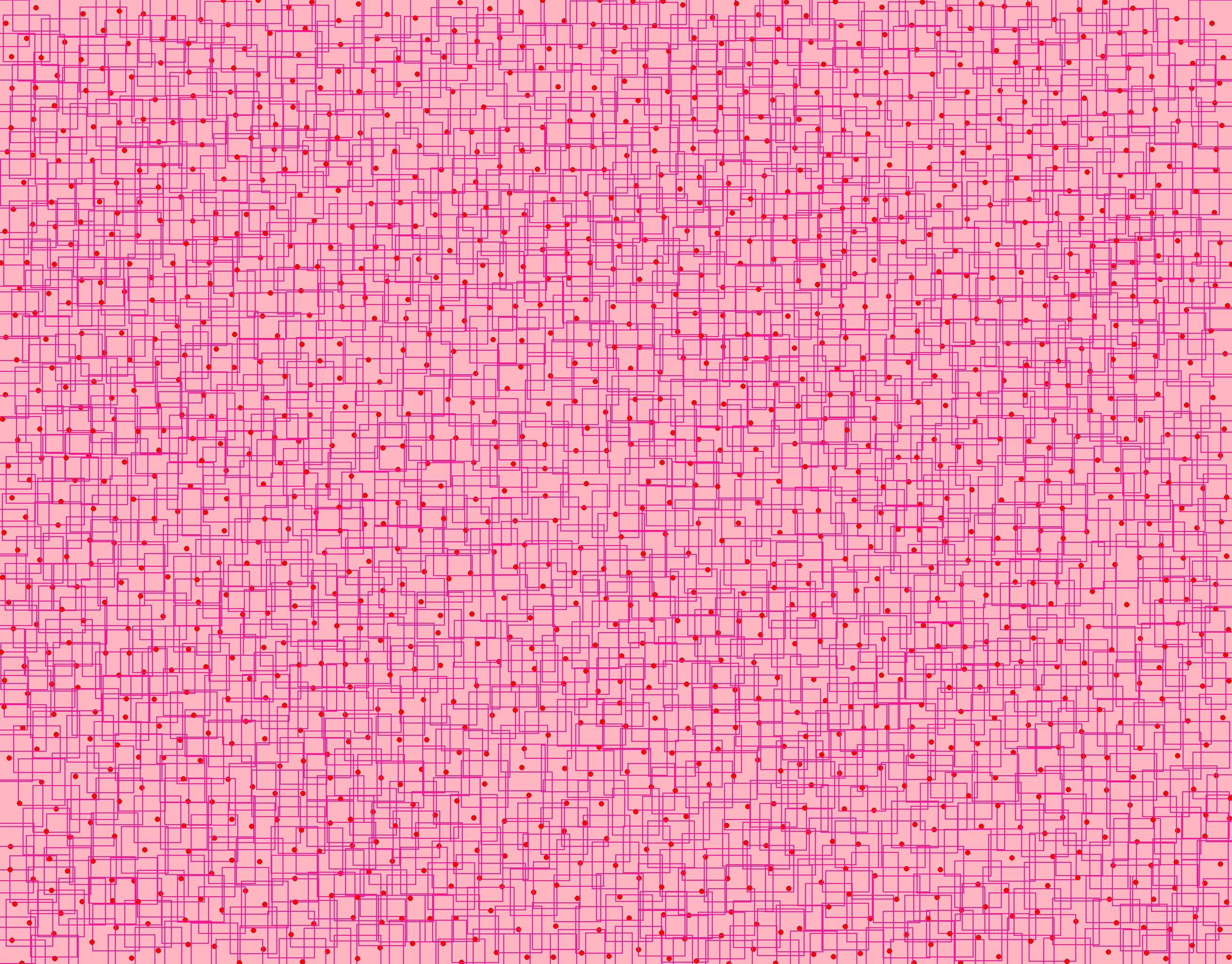}
    \end{subfigure}\hfill
    \begin{subfigure}{0.235\textwidth}
        \includegraphics[width=\linewidth]{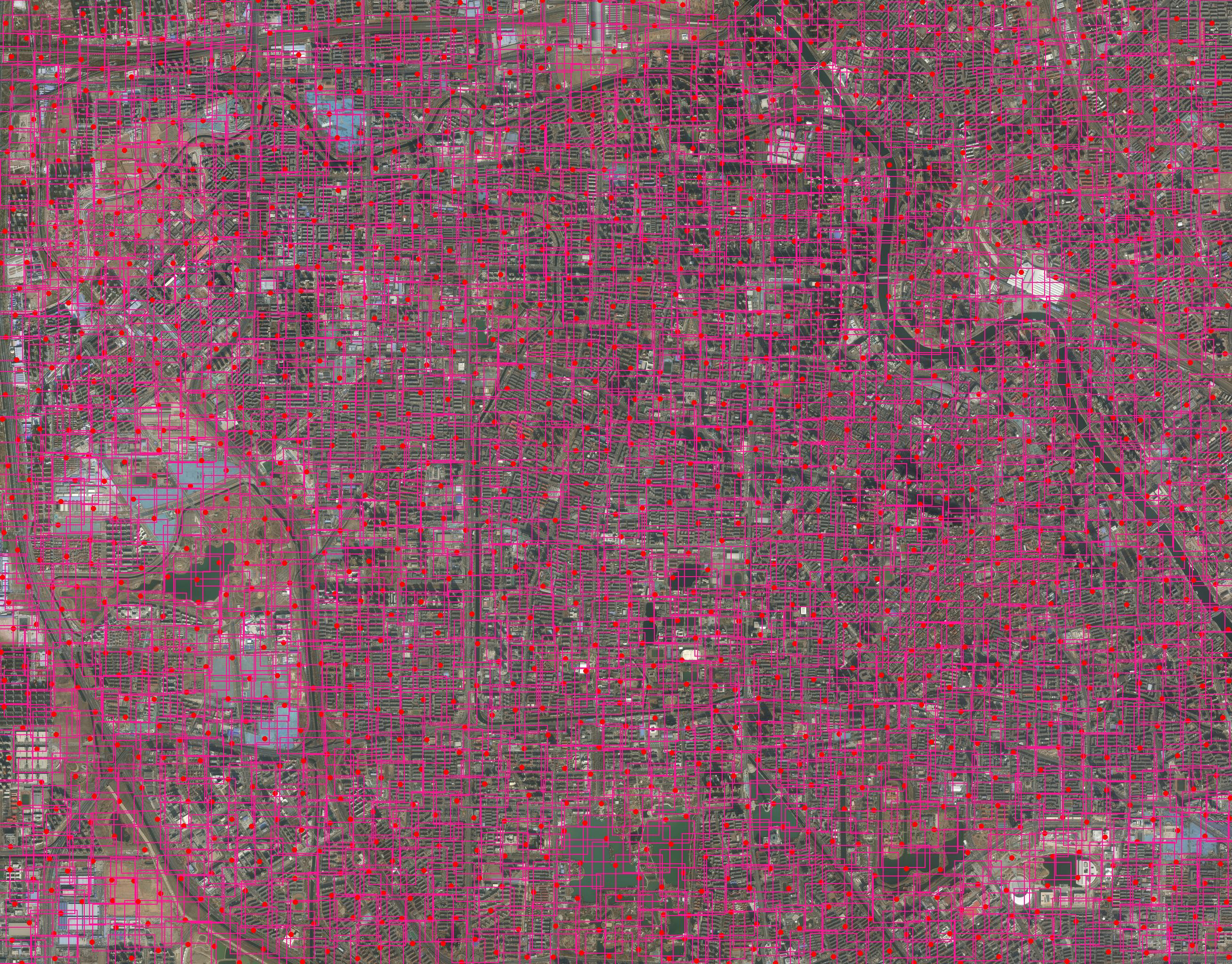}
    \end{subfigure}

    \caption{Sampling with Poisson Disk, the red points represent the sampling points generated using the Poisson Disk method, while the pink box indicates the area extracted based on these Poisson Disk sampling points.}
\end{figure}

We employed the Enhanced Intersection over Union (EIoU) to refine bounding box predictions, effectively addressing the limitations of traditional Intersection over Union (IoU) by incorporating geometric information. The EIoU loss function is defined as follows\cite{eiou,ciou}:

\[ \text{EIoU} = 1 - \text{IoU} + \rho^2 + v \]

\[ \text{IoU} = \frac{\text{Area of Overlap}}{\text{Area of Union}} \]

In this equation, \( \rho^2 \) denotes the center point distance loss, while \( v \) represents the aspect ratio loss. This approach significantly enhances both the accuracy and efficiency of the model.

Subsequently, these results were utilized as synthetic data and mixed with the original dataset.

\section{Experiments}

\subsection{Initial training}

We utilized the XView dataset as our initial dataset. XView is a large-scale aerial image dataset containing over 1 million objects annotated with bounding boxes and class labels, designed for advancing computer vision tasks in satellite imagery analysis\cite{xview1}. To align with the test image dimensions, we cropped the images from the XView dataset into smaller images of size 640×640 and remapped the label positions accordingly. 

Initial training was conducted on the XView dataset using both YOLOv11 and MobileNetV3-SSD. 

% \begin{figure}[htbp]
%     \centering
%     \begin{minipage}[t]{0.45\linewidth}
%         \centering
%         \includegraphics[width=0.9\linewidth]{figure/1111/d_1.jpg}
%     \end{minipage}
%     \begin{minipage}[t]{0.45\linewidth}
%         \centering
%         \includegraphics[width=0.9\linewidth]{figure/1111/d_2.jpg}
%     \end{minipage}
%     \caption{YOLO Small image detection results}
%     \label{fig:image_group}
%  \end{figure}

\subsection{metrics}
We systematically evaluate new results by traversing all outcomes in the original dataset, employing the Intersection over Union (IoU) metric to assess accuracy. :

In this context, the Area of Overlap refers to the region where the detection box intersects with the ground truth box, while the Area of Union represents the total area covered by both boxes, excluding any overlapping regions. A result is considered accurate if its IoU value exceeds 50%.

Subsequently, we evaluate model performance using several metrics: accuracy, precision, recall, and F1-score:

\[ \text{Accuracy} = \frac{\text{TP} + \text{TN}}{\text{TP} + \text{FP} + \text{FN} + \text{TN}} \]

\[ \text{Precision} = \frac{\text{TP}}{\text{TP} + \text{FP}}\hspace{1cm} 
 \text{Recall} = \frac{\text{TP}}{\text{TP} + \text{FN}}\]
% \[ \text{Recall} = \frac{\text{TP}}{\text{TP} + \text{FN}}  \]

\[ \mathrm {F1 Score}=2\times\frac {\mathrm {Precision}\times\mathrm {Recall}}{{\mathrm {Precision}+ \mathrm {Recall } }}   \] 

Where:

TP denotes the number of instances accurately identified as positive by the model.

TN indicates the number of instances correctly classified as negative by the model.

FP signifies instances incorrectly labeled as positive by the model.

FN represents instances inaccurately categorized as negative by the model.

Furthermore, we calculate the Mean Average Precision (mAP):

\[ \mathrm {mAP}= \dfrac {1}{N}\sum_{i=1}^{N}\mathrm {AP}_i   \] 

Here \(N\) denotes the total number of categories and \(AP_i\) reflects average precision for category \(i.\)

\subsection{Comparisons}

\begin{table*}[tbp]
  \centering
    \begin{tabular}{cc|ccc|ccc|ccc|ccc}
        &     & \multicolumn{3}{c|}{\textbf{Tianjin}} & \multicolumn{3}{c|}{\textbf{Shanghai}} & \multicolumn{3}{c|}{\textbf{Xiamen}} & \multicolumn{3}{c}{\textbf{Average}} \\
        &     & Accuracy & F1 & mAP & Accuracy & F1 & mAP & Acc & F1 & mAP & Accuracy & F1 & mAP \\
    \midrule
    \multirow{2}[2]{*}{\textbf{RetinaNet}} & whole & 0.039 & 0.003 & 0.01 & 0.042 & 0.008 & 0.02 & 0.035 & 0.007 & 0.02 & 0.039 & 0.006 & 0.02 \\
        & 640-cut & 0.603 & 0.640 & 0.52 & 0.605 & 0.635 & 0.53 & 0.598 & 0.644 & 0.54 & 0.602 & 0.640 & 0.53 \\
    \midrule
    \multirow{2}[2]{*}{\textbf{SSD\_VGG}} & whole & 0.063 & 0.013 & 0.02 & 0.066 & 0.017 & 0.03 & 0.067 & 0.01 & 0.01 & 0.065 & 0.013 & 0.02 \\
        & 640-cut & 0.693 & 0.521 & 0.47 & 0.691 & 0.518 & 0.46 & 0.69 & 0.525 & 0.48 & 0.691 & 0.521 & 0.47 \\
    \midrule
    \multirow{2}[2]{*}{\textbf{Faster R-CNN}} & whole & 0.046 & 0.004 & 0.01 & 0.049 & 0.007 & 0.02 & 0.05 & 0.002 & 0 & 0.048 & 0.004 & 0.01 \\
        & 640-cut & 0.560 & 0.590 & 0.51 & 0.562 & 0.597 & 0.47 & 0.555 & 0.588 & 0.49 & 0.559 & 0.592 & 0.49 \\
    \midrule
    \multirow{2}[2]{*}{\textbf{YOLOv11}} & whole & 0.082 & 0.002 & 0.01 & 0.085 & 0.005 & 0.02 & 0.079 & 0.001 & 0.03 & 0.082 & 0.003 & 0.02 \\
        & 640-cut & 0.619 & 0.666 & 0.50 & 0.616 & 0.661 & 0.49 & \cellcolor[rgb]{ 1,  1,  0}0.794 & 0.722 & \cellcolor[rgb]{ 1,  1,  0}0.69 & 0.676 & 0.683 & 0.56 \\
    \midrule
    \multirow{2}[2]{*}{\textbf{MobileNetV3-SSD}} & whole & 0.076 & 0.013 & 0.33 & 0.079 & 0.01 & 0.30 & 0.073 & 0.012 & 0.32 & 0.076 & 0.012 & 0.32 \\
        & 640-cut & 0.700 & 0.668 & 0.65 & 0.702 & 0.672 & 0.62 & 0.595 & 0.665 & 0.63 & 0.666 & 0.668 & 0.63 \\
    \midrule
    \multirow{3}[1]{*}{\textbf{LRSAA (ours)}} & 1280-cut & 0.706 & 0.706 & 0.64 & 0.704 & 0.703 & 0.62 & 0.711 & 0.709 & 0.66 & 0.707 & 0.706 & 0.64 \\
        & 640-cut & \cellcolor[rgb]{ 1,  1,  0}0.796 & \cellcolor[rgb]{ 1,  1,  0}0.798 & 0.71 & 0.795 & \cellcolor[rgb]{ 1,  1,  0}0.802 & 0.70 & 0.781 & \cellcolor[rgb]{ 1,  1,  0}0.731 & \cellcolor[rgb]{ 1,  1,  0}0.69 & \cellcolor[rgb]{ 1,  1,  0}0.791 & \cellcolor[rgb]{ 1,  1,  0}0.777 & 0.70 \\
        & 320-cut & 0.794 & 0.789 & \cellcolor[rgb]{ 1,  1,  0}0.73 & \cellcolor[rgb]{ 1,  1,  0}0.798 & 0.785 & \cellcolor[rgb]{ 1,  1,  0}0.72 & 0.698 & 0.663 & 0.72 & 0.763 & 0.746 & \cellcolor[rgb]{ 1,  1,  0}0.72 \\
    \end{tabular}%
  \caption{Comparative Results: All algorithms were assessed using confidence boxes with a probability threshold exceeding 0.25, based on manually annotated urban remote sensing images with a resolution of 0.6 meters. Each city was represented by 20 annotated regions, each measuring 6400x6400 pixels.}
  \label{tab:addlabel}%
\end{table*}%

This study conducted experimental validation utilizing a self-constructed high-resolution remote sensing image dataset encompassing the entire urban areas of Shanghai, Tianjin, and Xiamen. Acknowledging the performance variations of object detection models across different image cropping scales, this research specifically selected three standard cropping sizes: 320×320 pixels, 640×640 pixels, and 1280×1280 pixels to systematically evaluate the robustness and adaptability of the proposed method.

To further investigate the model's generalizability and to compare the advantages and disadvantages of various architectures, three advanced object detection frameworks—RetinaNet\cite{retinanet}, SSD\_VGG\cite{ssd}, Faster R-CNN\cite{rcnn3}—were chosen for comprehensive evaluation. Specifically, not only was the detection performance of each model analyzed on complete original images but their performance was also reassessed after applying Poisson disk sampling cropping techniques to examine consistency and stability under diverse scenarios and conditions.

\subsection{Comparing Result}

The results of the comparison are detailed in Table I and Table III located in the appendix.

From the results presented in Table 1 and Table 2, it is evident that RetinaNet exhibits suboptimal performance when evaluated on whole images; however, its performance significantly improves with the use of 640-cut images. For instance, in Tianjin, the Accuracy increases from 0.039 to 0.603, while the F1-score rises from 0.003 to 0.640. Similarly, both SSD\_VGG and Faster R-CNN demonstrate enhanced performance with smaller image sizes. In Tianjin, SSD\_VGG's Accuracy improves from 0.063 to 0.693, whereas Faster R-CNN's Accuracy increases from 0.046 to 0.560. These trends are consistent across all cities analyzed, indicating that reduced image sizes contribute positively to the performance of these models.

YOLOv11 and MobileNetV3-SSD also show notable improvements with 640-cut images. YOLOv11's performance in Xiamen jumps from an Accuracy of 0.079 to 0.794 and an F1-score from 0.001 to 0.722. MobileNetV3-SSD demonstrates strong performance even with whole images but further improves with 640-cut images. For instance, in Tianjin, Accuracy increases from 0.076 to 0.700, and F1-score from 0.013 to 0.668. Both models maintain high performance across all cities, making them robust choices for object detection tasks.

The proposed LRSAA model outperforms all other models, especially with 640-cut and 320-cut images. In Tianjin, Accuracy reaches 0.796 with 640-cut images, and F1-score reaches 0.798. The model consistently achieves the highest scores across all metrics and cities, demonstrating its superiority in object detection tasks. 

In summary, while each model's performance varies across different datasets, LRSAA consistently showcases advantages in multiple evaluation metrics, especially when addressing smaller datasets. This indicates not only improved accuracy in object detection but also excellent performance under resource-constrained conditions, providing robust support for practical applications.

\subsection{Synthetic Data Training and Evaluation}

To enhance the generalization and robustness of the models, we conducted random sampling on a large-scale dataset of remotely sensed annotated images from Tianjin, resulting in the generation of a series of synthetic data samples with dimensions of 640×640 pixels. These synthetic data were subsequently integrated into the original dataset at a specified ratio, thereby forming an extended dataset. Utilizing this extended dataset, we retrained LRSAA.

\begin{table}[H]
  \centering
  %\caption{Synthetic data training result}
    \begin{tabular}{cc|ccc}
    Original Data & Composite data & Acurracy & F1-score & mAP \\
    \midrule
    0\% & 100\% & 0.707 & 0.706 & 0.64 \\
    20\% & 80\% & 0.712 & 0.720 & 0.65 \\
    40\% & 60\% & 0.710 & 0.698 & 0.64 \\
    60\% & 40\% & \cellcolor[rgb]{ 1,  1,  0}0.718 & 0.733 & \cellcolor[rgb]{ 1,  1,  0}0.67 \\
    80\% & 20\% & 0.716 & \cellcolor[rgb]{ 1,  1,  0}0.739 & 0.66 \\
    \end{tabular}%
  \caption{Synthetic data training result}
  \label{tab:addlabel}%
\end{table}%

Upon completion of model training, we performed detailed evaluations to assess the practical application effectiveness of the improved models using large-scale manually annotated remotely sensed data from Tianjin, Shanghai and Xiamen. The evaluation results are summarized as follows:

The analysis results indicate a significant trend toward performance improvement in object detection and prediction as the proportion of synthetic data increases, evident across all metrics including Accuracy, F1-score, and mAP.

\section{Conclusion and Future Work}

\subsection{Conclusion}

In this paper, we present a novel framework for large-scale remote sensing image target recognition and automatic annotation, referred to as LRSAA. This approach integrates the YOLOv11 and MobileNetV3-SSD object detection algorithms through ensemble learning, thereby enhancing model performance while simultaneously reducing computational resource requirements. The implementation of Poisson disk sampling segmentation along with the EIOU-NMS optimizes both training and inference processes, achieving a commendable balance between accuracy and speed. 

The LRSAA model was trained and evaluated using the XView dataset and subsequently applied to remote sensing images from Tianjin, Shanghai and Xiamen. Notably, significant improvements in recognition capabilities were observed when synthetic data was incorporated into the training process. This work contributes to advancing remote sensing image analysis by enabling more efficient and accurate acquisition and utilization of geographic information.

\subsection{Future Work}

Looking ahead, several avenues for future research can be explored to further develop the LRSAA framework. First, integrating more advanced deep learning models and algorithms has the potential to significantly enhance both the accuracy and efficiency of object detection in remote sensing images. Investigating the capabilities of transformer-based models or other state-of-the-art architectures may lead to additional performance improvements. Given that our ensemble learning approach integrates the output results from each model, it facilitates the incorporation of new algorithms with relative ease.

Second, examining the computational efficiency of the LRSAA framework across various hardware platforms—such as edge devices and cloud computing environments—could provide valuable insights into its scalability and practicality for large-scale applications.

Lastly, exploring real-time object detection capabilities within the LRSAA framework is crucial for time-sensitive applications such as disaster monitoring and dynamic environmental tracking. Optimizing the model for real-time performance without compromising accuracy represents a challenging yet essential direction for future work.

\newpage
\appendix
\section{Appendix of Detection result comparisons}
\begin{minipage}{\textwidth}

  \begin{table}[H] % 使用[H]位置选项来尽量保持当前位置
    \centering
    \begin{tabular}{m{3cm} m{0.25\textwidth} m{0.25\textwidth} m{0.25\textwidth}}
      & \multicolumn{1}{c}{\textbf{Tianjin}} & \multicolumn{1}{c}{\textbf{Shanghai}} & \multicolumn{1}{c}{\textbf{Xiamen}} \\
      \midrule
      \textbf{Original} & \includegraphics[width=0.23\textwidth]{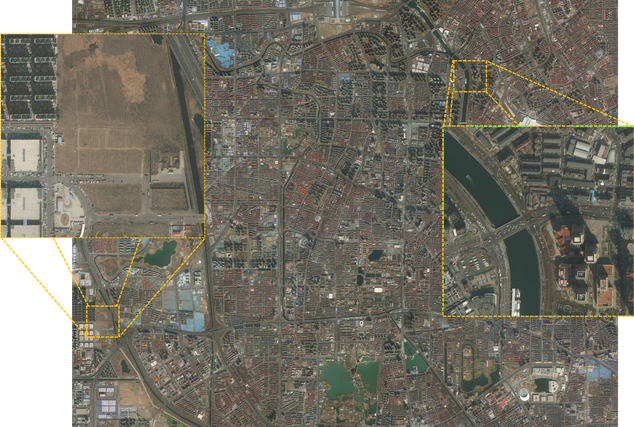} & \includegraphics[width=0.25\textwidth]{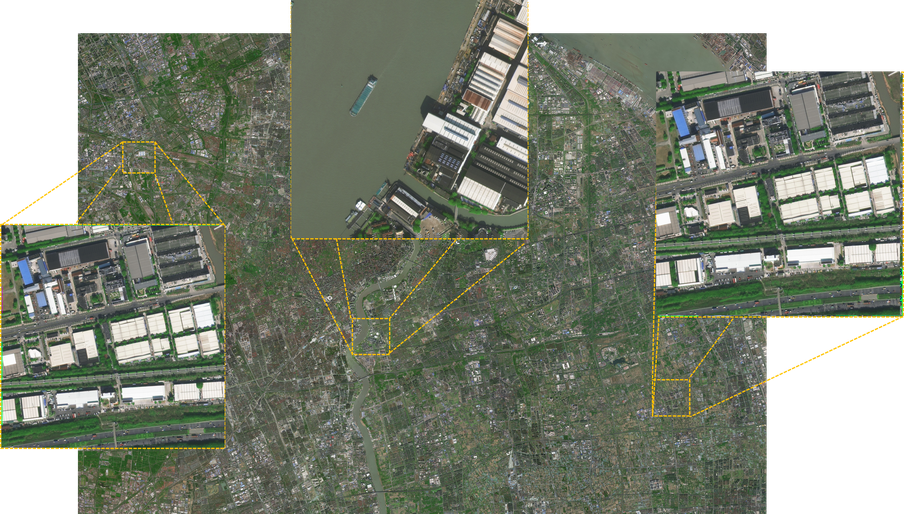} & \includegraphics[width=0.2\textwidth]{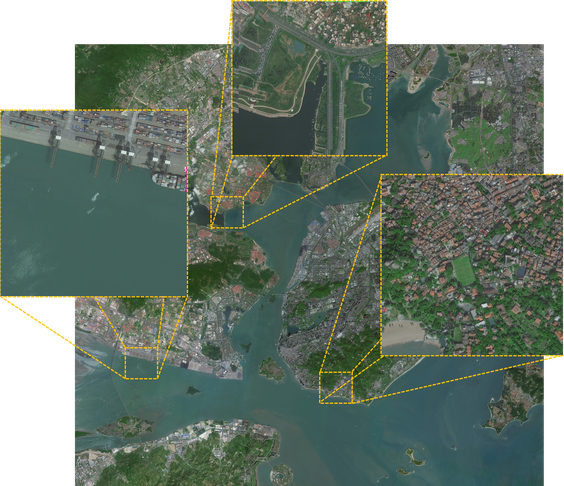} \\
      \textbf{RetinaNet} & \includegraphics[width=0.23\textwidth]{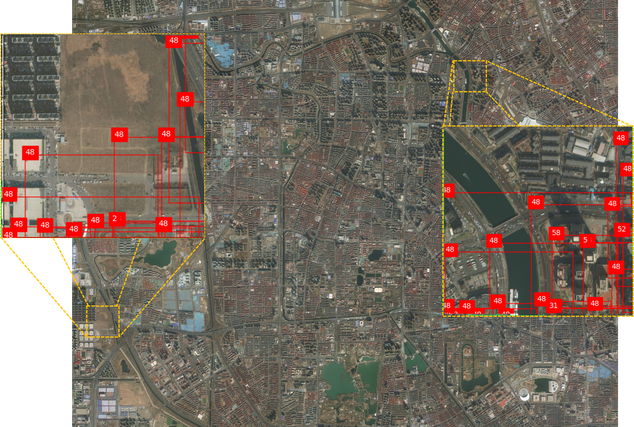} & \includegraphics[width=0.25\textwidth]{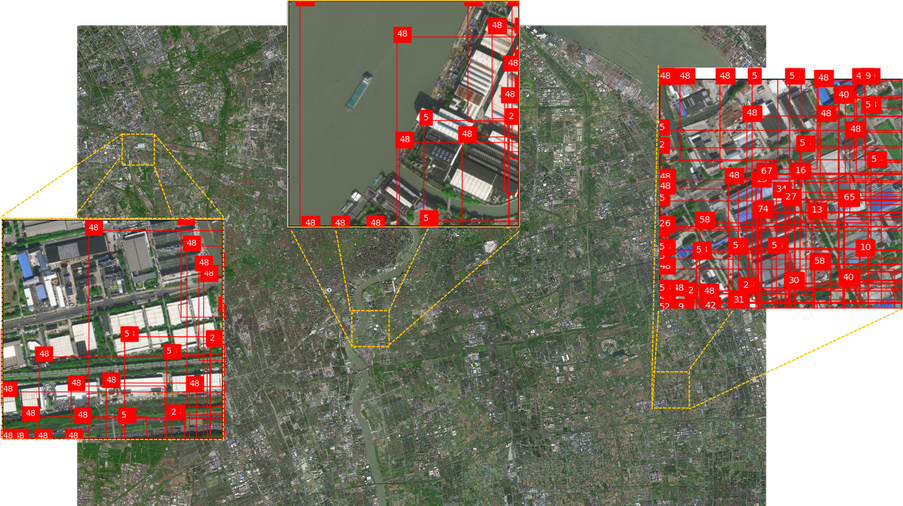} & \includegraphics[width=0.2\textwidth]{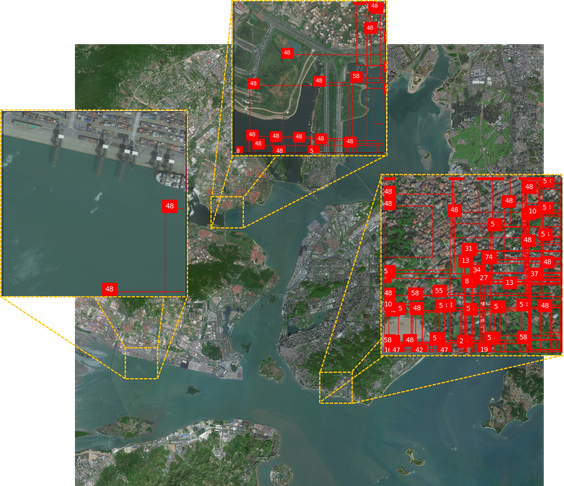} \\
      \textbf{SSD\_VGG} & \includegraphics[width=0.23\textwidth]{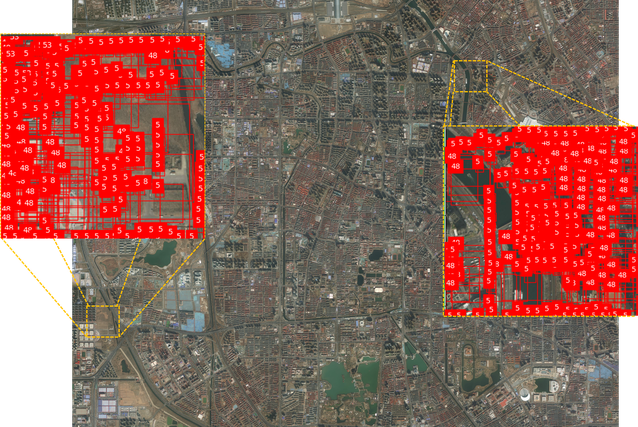} & \includegraphics[width=0.25\textwidth]{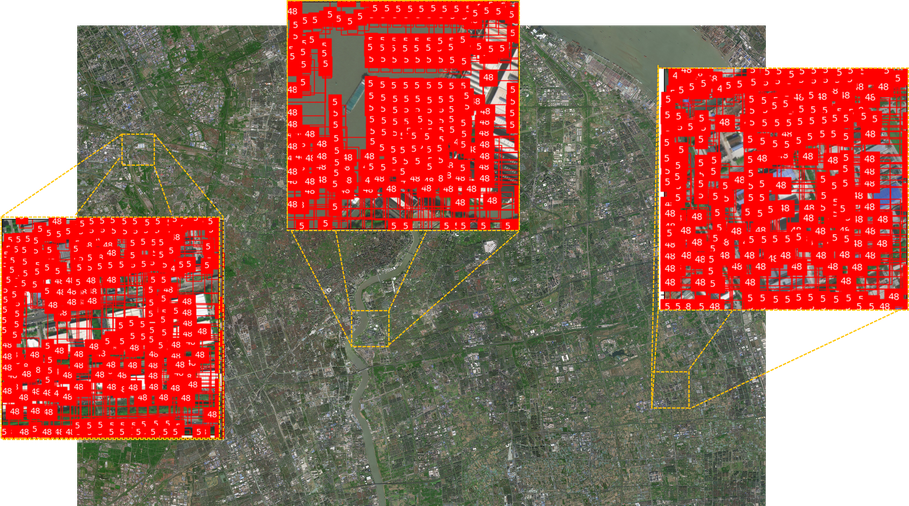} & \includegraphics[width=0.2\textwidth]{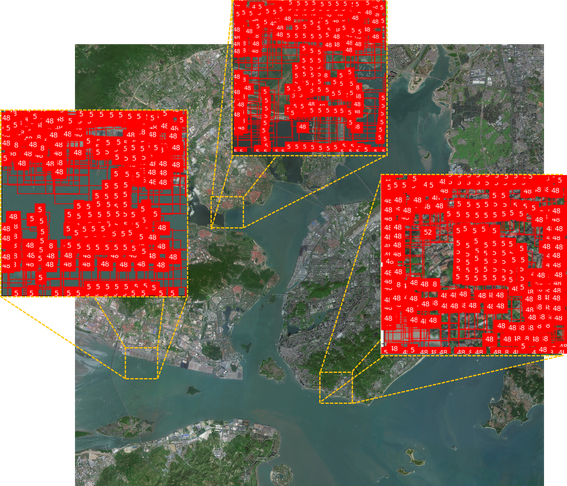} \\
      \textbf{Faster R-CNN} & \includegraphics[width=0.23\textwidth]{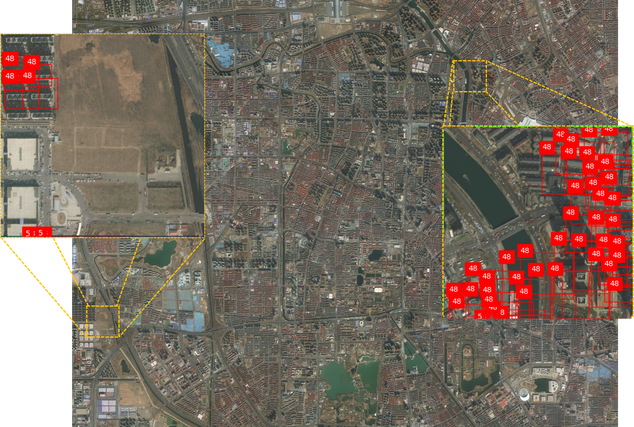} & \includegraphics[width=0.25\textwidth]{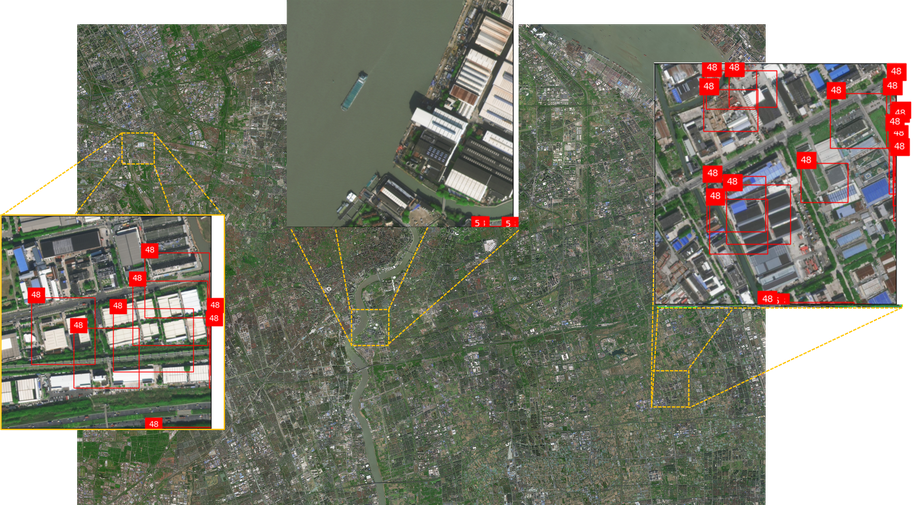} & \includegraphics[width=0.2\textwidth]{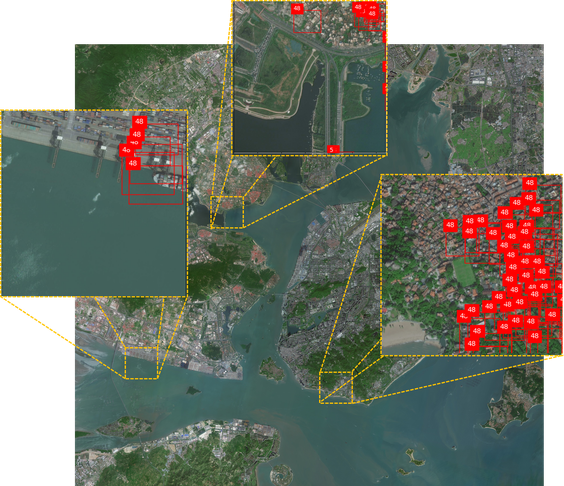} \\
      \textbf{YOLOv11} & \includegraphics[width=0.23\textwidth]{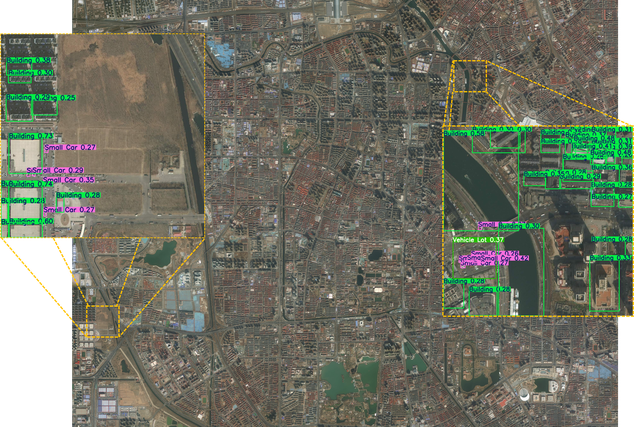} & \includegraphics[width=0.25\textwidth]{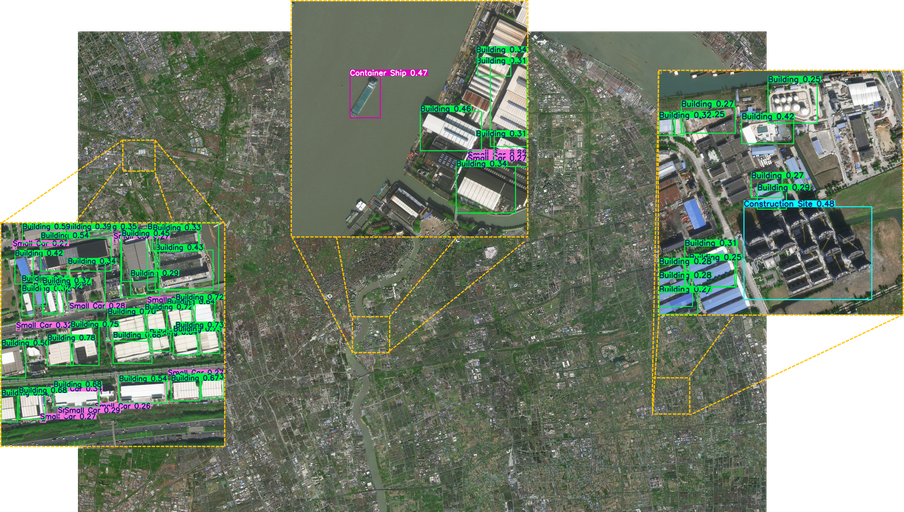} & \includegraphics[width=0.2\textwidth]{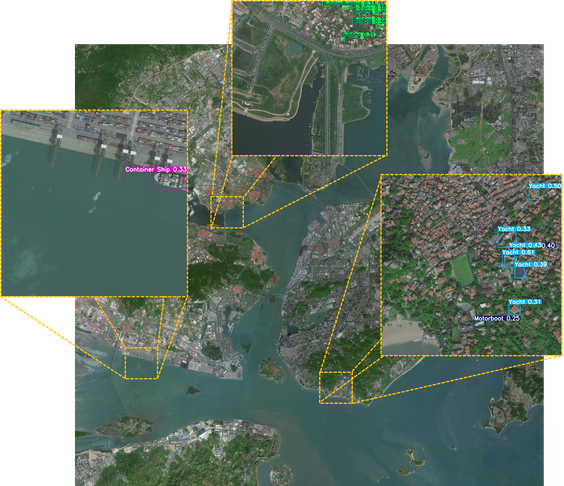} \\
      \textbf{MobileNetV3-SSD} & \includegraphics[width=0.23\textwidth]{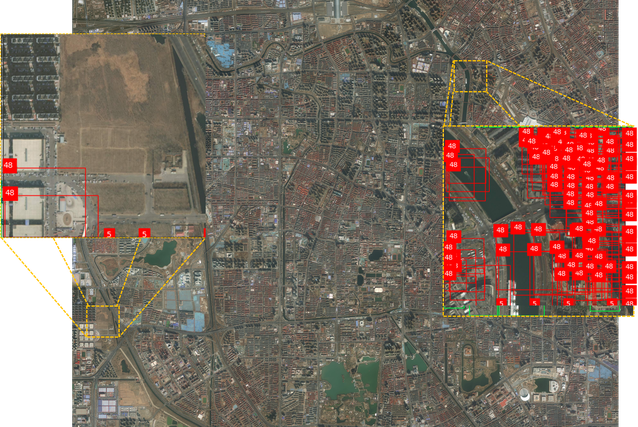} & \includegraphics[width=0.25\textwidth]{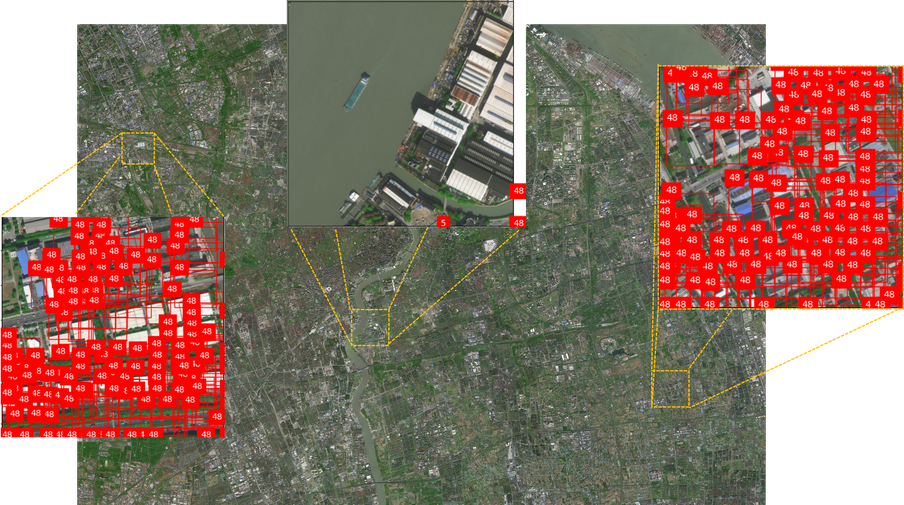} & \includegraphics[width=0.2\textwidth]{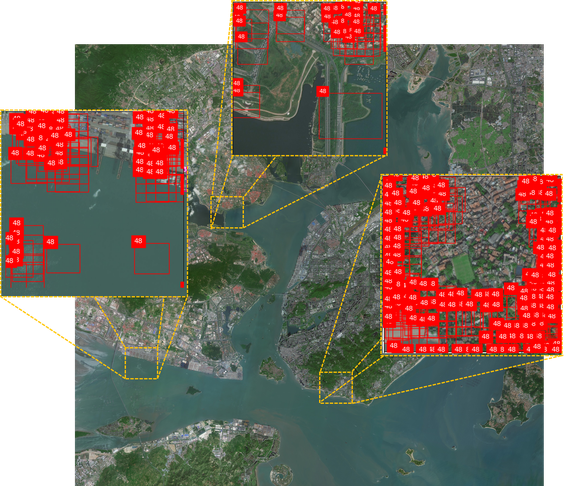} \\
      \textbf{LRSAA (ours)} & \includegraphics[width=0.23\textwidth]{figure/tianjinend/lrsaa.png} & \includegraphics[width=0.25\textwidth]{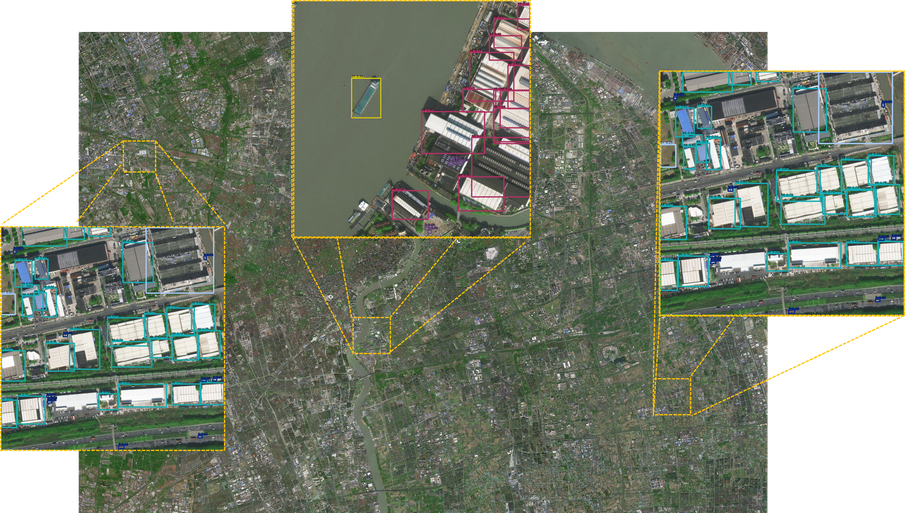} & \includegraphics[width=0.2\textwidth]{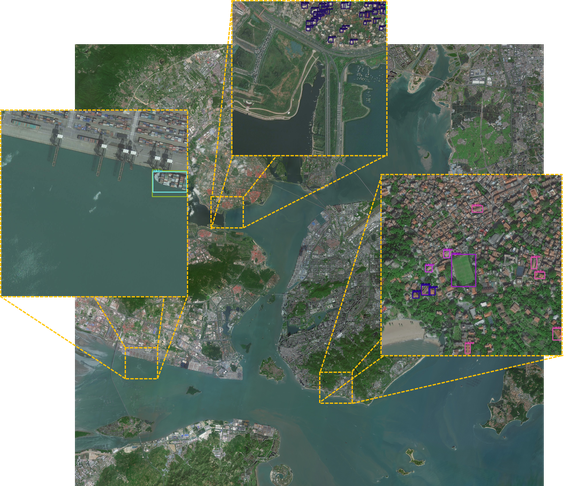} \\
    \end{tabular}%
    \caption{Detection comparisons: Complete ground truth targets are delineated with a uniform red box, whereas the results presented in the smaller images utilize a variety of colors to represent both the outcomes and their associated probabilities.}
    \label{tab:addlabel}
  \end{table}
  
\end{minipage}

\end{document}